\documentclass[11pt]{article}

\usepackage[preprint]{acl}

\usepackage{times}
\usepackage{latexsym}

\usepackage[T1]{fontenc}

\usepackage[utf8]{inputenc}

\usepackage{microtype}

\usepackage{inconsolata}

\usepackage{xurl}

\usepackage{graphicx}

\usepackage{amsmath}
\usepackage{amssymb}
\usepackage{booktabs}
\usepackage{stfloats}
\usepackage{multirow}
\usepackage{xcolor}
\usepackage{xspace}

\newcommand{\daps}{DAPS\xspace}
\newcommand{\grpo}{GRPO\xspace}
\newcommand{\grpox}{\mbox{GRPO-X}\xspace}
\newcommand{\ppo}{PPO\xspace}
\newcommand{\ourmethod}{S3AC\xspace}
\newcommand{\rpp}{REINFORCE++\xspace}
\newcommand{\flash}{FlashREINFORCE\xspace}
\newcommand{\qwensmall}{Qwen2.5-Math-1.5B\xspace}
\newcommand{\rd}{R1-Distill-1.5B\xspace}
\newcommand{\smid}{Intern-S2-Preview-35B\xspace}
\newcommand{\sbig}{Intern-S2-397B\xspace}

\title{S3AC: Stabilized, Selective, and Scalable Actor--Critic for LLM Reasoning}

\author{
  Shuo Shen\textsuperscript{1}, Guangran Cheng\textsuperscript{1}, Chengqi Lyu\textsuperscript{1},\\
  \bfseries Xiangsen Wang\textsuperscript{2}, Yanan Sun\textsuperscript{1,*}, Wenwei Zhang\textsuperscript{1,*}\\
  \textsuperscript{1}Shanghai AI Laboratory \quad \textsuperscript{2}Independent Researcher
}

\begin{document}
\maketitle
\begingroup
\renewcommand{\thefootnote}{\fnsymbol{footnote}}
\footnotetext[1]{Corresponding authors.}
\endgroup

\begin{abstract}
Critic-free methods are widely used in reinforcement learning with verifiable rewards (RLVR) for long chain-of-thought (CoT) reasoning, avoiding difficult and costly critic training.
Common alternatives lack token-level credit assignment and typically rely on grouped rollouts to estimate prompt-specific baselines.
A critic can address both limitations, but long-CoT critic training is hindered by insufficient coverage of trajectories with diverse lengths and early gradients dominated by the newly initialized value head.
We introduce S3AC (Stabilized, Selective, and Scalable Actor--Critic) for LLM reasoning, built around a practical critic-training recipe.
The trained critic provides token-specific advantages and prompt-specific baselines.
Using the prompt-level baseline to estimate difficulty, we propose Difficulty-Aware Prompt Sampling (\daps).
\daps combines this estimate with outcome history to select prompts before generation, reducing rollout waste without offline difficulty initialization.
S3AC improves group-rollout performance across dense and mixture-of-experts models from 1.5B to 397B parameters, outperforming a modern \grpo variant (\grpox).
At 35B, it sustains single-rollout training while tested critic-free baselines deteriorate after initial gains.
\end{abstract}

\begin{figure*}[t]
  \centering
  \includegraphics[width=\textwidth]{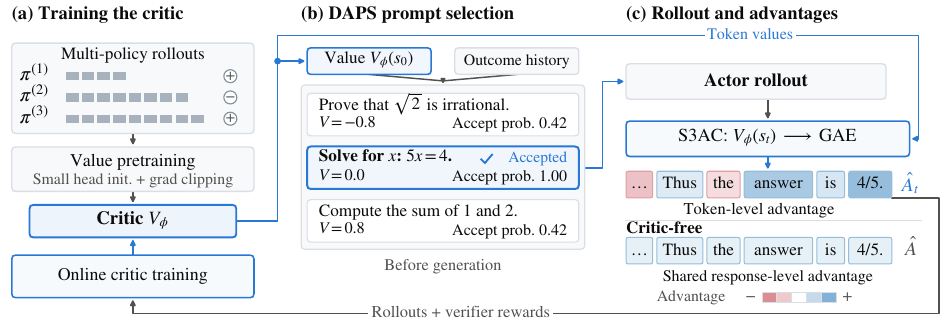}
  \vspace{-1em}
  \caption{Overview of \ourmethod for long-CoT RLVR. (a) Critic value pretraining and online updates (\S\ref{sec:critic}). (b) \daps selects prompts using critic values and outcome history (\S\ref{sec:daps}). (c) \ourmethod uses token-level GAE advantages; critic-free methods share one advantage across the response (\S\ref{sec:comparison}).}
  \label{fig:overview}
  \vspace{-1em}
\end{figure*}

\section{Introduction}\label{sec:intro}

Critic-free methods, including \grpo, are widely used in reinforcement learning with verifiable rewards (RLVR) for large language model (LLM) reasoning \citep{deepseek_v41flash_2026,kimi_k25_2026,kimi_k3_2026,xiaomi_mimo_v26_2026,intern_s2_preview_2026}.
They avoid the difficulty and cost of training a value model for long responses under sparse rewards and an evolving policy.
However, replacing the critic with reward statistics limits how learning signals distinguish prompts and reasoning steps.

A failure may be expected on a difficult prompt but surprising on an easy one, motivating a prompt-specific baseline.
\grpo estimates this baseline from grouped rollouts, requiring each group to wait for its slowest response \citep{xu_single-stream_2025,hou_single-rollout_2026}.
Common single-rollout alternatives, including \rpp~\citep{hu_reinforce_2025} and \flash~\citep{hu_flashreinforce_2026}, use batch statistics instead.
These shared baselines do not account for differences in prompt difficulty.

Even with a prompt-specific baseline, a single response-level advantage cannot distinguish the contributions of individual tokens.
When an answer is incorrect, useful intermediate reasoning can receive the same negative signal as steps that lead away from a solution \citep{wu_rethinking_2026}.
This scalar signal does not reflect how the difficulty of reaching a correct answer changes as reasoning progresses.

A learned critic offers a common route to addressing both limitations: a prompt-conditioned baseline without grouped rollouts and token-specific learning signals throughout the response \citep{sutton_policy_1999}.
However, learning reliable value estimates for long reasoning trajectories remains challenging.
Our analysis highlights two obstacles to effective value learning: initial-policy trajectories provide limited supervision on long, correct responses, while early gradients can be dominated by the newly initialized value head.

To address these obstacles, we introduce S3AC (Stabilized, Selective, and Scalable Actor--Critic) for LLM reasoning (Figure~\ref{fig:overview}).
We pretrain the critic on trajectories from policies of different strengths to broaden coverage across response lengths and outcomes.
Small-scale value-head initialization controls initial predictions, while a larger gradient-clipping threshold reduces excessive rescaling of backbone updates by head-dominated gradients.
Updated online, the critic provides difficulty-aware, token-specific advantages throughout the response through generalized advantage estimation \citep{schulman_high-dimensional_2016}.

The critic's usefulness extends to deciding what to generate.
Because its prompt-conditioned estimate is available before rollout, it can guide prompt selection as well as policy learning.
We propose Difficulty-Aware Prompt Sampling (\daps), which combines this estimate with outcome history to select prompts of intermediate estimated difficulty.
Some existing selectors require offline-initialized statistics or an auxiliary value model \citep{xu_single-stream_2025,gao_prompt_2025}; \daps reuses the critic already trained for policy learning.

We evaluate two dense and two multimodal mixture-of-experts models spanning 1.5B to 397B parameters.
Under group rollout, actor--critic training improves average performance across all four models, outperforming \grpox on all seven evaluated benchmarks at 35B and all five at 397B.
At 35B, it sustains single-rollout performance while the tested critic-free baselines decline after initial gains (Figure~\ref{fig:bench-35b-average}).
\daps further improves both rollout settings over uniform prompt sampling and reduces trajectories discarded by group filtering.

\section{Related Work}\label{sec:related}

\paragraph{Critic-based RL.}
\ppo \citep{schulman_proximal_2017} uses a learned critic with GAE \citep{schulman_high-dimensional_2016} to estimate advantages for policy optimization.
For long-CoT reasoning, VC-PPO \citep{yuan_whats_2025} pretrains the value model and decouples the GAE parameters of the actor and critic, while VAPO \citep{yue_vapo_2025} adapts $\lambda$ to response length.
VinePPO \citep{kazemnejad_vineppo_2025}, EVPO \citep{pan_evpo_2026}, and AsyPPO \citep{liu_asymmetric_2025} improve value estimates through Monte Carlo rollouts, explained-variance gating, and critic ensembles, respectively.
SAO \citep{hou_single-rollout_2026} and CompactionRL \citep{li_compactionrl_2026} apply \ppo to single-rollout training of large MoE models.
While these methods address different aspects of actor--critic training, our work brings critic initialization, online value learning, and critic-guided prompt selection together in a unified framework.
Our study combines targeted diagnostics and ablations with evaluations across rollout settings and model scales, including group-rollout training at 397B parameters.

\paragraph{Critic-free RL.}
Group-based methods such as \grpo \citep{shao_deepseekmath_2024}, RLOO \citep{ahmadian_back_2024}, and DAPO \citep{yu_dapo_2025} derive baselines from multiple responses to the same prompt.
Single-rollout methods use batch statistics (\rpp; \citealp{hu_reinforce_2025}), add a sequence-level trust region to a batch-mean baseline (\flash; \citealp{hu_flashreinforce_2026}), or track per-prompt values with a tabular estimator (SPO; \citealp{xu_single-stream_2025}).
These methods typically construct advantages from sequence-level reward statistics without learning a value function over intermediate response prefixes.

\paragraph{Difficulty-aware prompt sampling.}
Dynamic sampling \citep{yu_dapo_2025} requires multiple responses per prompt and discards responses from uninformative groups after generation.
Other methods estimate difficulty before rollout from reward history (GRESO; \citealp{zheng_act_2025}), from a tracker initialized with offline rollouts (SPO), or from a separately trained prompt-level value model (PCL; \citealp{gao_prompt_2025}).
\daps combines pass-rate history with value estimates from the existing \ourmethod critic, avoiding a separately trained difficulty model and supporting both group and single rollout.

\section{Method}\label{sec:method}

\ourmethod is an actor--critic framework for long-CoT RLVR that trains a value model for both policy learning and prompt selection (Figure~\ref{fig:overview}).
The critic provides prompt-conditioned baselines and token-specific advantages, while its initial-state value guides prompt sampling before generation.
We first formulate the actor--critic update (\S\ref{sec:prelim}), then examine the critic's role under group and single rollout (\S\ref{sec:comparison}), describe its training recipe (\S\ref{sec:critic}), and introduce \daps (\S\ref{sec:daps}).

\subsection{Actor--Critic Formulation}\label{sec:prelim}

We model response generation as a token-level MDP.
Given a prompt $x$, the actor $\pi_\theta$ generates a response token by token: the state $s_t$ contains $x$ and the first $t$ response tokens, and the action $a_t$ is the next token.
A verifier gives a scalar reward $R\in\{-1,+1\}$ on the last token only, and we use $\gamma=1$.
Following the actor--critic formulation \citep{sutton_policy_1999}, a learned critic $V_\phi(s_t)$ approximates the expected return $\mathbb{E}_{\pi_\theta}[R\mid s_t]$.
At $s_0$, the estimate depends on the prompt alone; subsequent values also condition on the generated response prefix.

For a response of $T$ tokens, we compute token-specific advantages using GAE \citep{schulman_high-dimensional_2016}:
\begin{equation}
  \hat A_t=\sum_{l=0}^{T-t-1}\lambda^{l}\delta_{t+l},
  \label{eq:gae}
\end{equation}
where $\delta_t=R_t+V_\phi(s_{t+1})-V_\phi(s_t)$ is the TD error, with $R_t=R$ at the last token and $0$ elsewhere.
The terminal value is $V_\phi(s_T)=0$.
Equivalently, $\hat A_t=G^\lambda_t-V_\phi(s_t)$, where the $\lambda$-return $G^\lambda_t$ bootstraps from later values when $\lambda<1$, trading variance for bias.

For actor updates, we adopt CISPO \citep{minimax_minimax-m1_2025}, whose core surrogate is
\begin{equation}
  \begin{aligned}
    \mathcal{J}_{\mathrm{actor}}(\theta)
      &=\hat{\mathbb{E}}_t\!\left[
        \operatorname{sg}(\bar r_t\hat A_t)
        \log\pi_\theta(a_t\mid s_t)\right],\\
    \bar r_t&=\operatorname{clip}(r_t,1-\epsilon_{\mathrm{low}},
                                      1+\epsilon_{\mathrm{high}}),
  \end{aligned}
  \label{eq:cispo}
\end{equation}
where $r_t=\pi_\theta(a_t\mid s_t)/\pi_{\theta_{\mathrm{old}}}(a_t\mid s_t)$ is the importance ratio relative to the actor before the current batch of updates, and $\operatorname{sg}$ denotes stop-gradient.
The clipped importance weights and advantages are thus treated as constants when differentiating the actor objective.
To address training--inference mismatch, we combine importance-weight corrections with masking at both the token and sequence levels (\S\ref{sec:setup}; Appendix~\ref{app:config}).

For critic updates, we regress token-level values toward the Monte Carlo return $R$ using a clipped squared-error loss \citep{engstrom_implementation_2020}.
Let $V_t=V_\phi(s_t)$ and $V_t^{\mathrm{old}}=V_{\phi_{\mathrm{old}}}(s_t)$.
The objective is
\begin{equation}
  \mathcal L_{\mathrm{critic}}
  =\tfrac12\hat{\mathbb E}_t
  \left[\max\left\{
  (V_t-R)^2,\,
  (\bar V_t-R)^2
  \right\}\right],
  \label{eq:critic-loss}
\end{equation}
where $\bar V_t$ is $V_t$ clipped to $[V_t^{\mathrm{old}}-\epsilon_v,\;V_t^{\mathrm{old}}+\epsilon_v]$, and $\epsilon_v$ controls value clipping.
The expectation averages over valid response tokens.

\subsection{How the Critic Shapes Learning Signals}\label{sec:comparison}

\begin{figure}[t]
  \centering
  \includegraphics[width=\columnwidth]{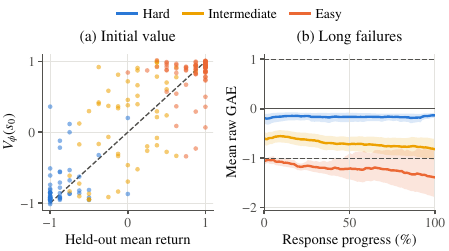}
  \caption{Prompt-conditioned values and advantages (35B single rollout). (a) Initial value versus held-out return. (b) Long failures ($>20$K tokens), using length-adaptive GAE with $\alpha=1.5$.}
  \label{fig:per-prompt-baseline}
\end{figure}

We examine how the critic shapes learning signals across prompts of different difficulty and across positions within a response.
The initial critic value $V_\phi(s_0)$ estimates the policy's expected return for a prompt before generation.
Figure~\ref{fig:per-prompt-baseline}a plots these values against mean Monte Carlo (MC) returns from independent rollouts, using a frozen 35B actor--critic pair with one point per prompt.
The values track differences in empirical returns, supporting their use as prompt-specific baselines.
The prompt-specific baseline is not limited to the initial position: subsequent values remain conditioned on the prompt throughout the response.
Figure~\ref{fig:per-prompt-baseline}b illustrates this distinction in long failed responses: although all receive the same terminal reward of $-1$, their mean raw advantages remain close to zero for hard prompts and below $-1$ for easy prompts across response progress.

\begin{figure}[t]
\centering
\includegraphics[width=\columnwidth]{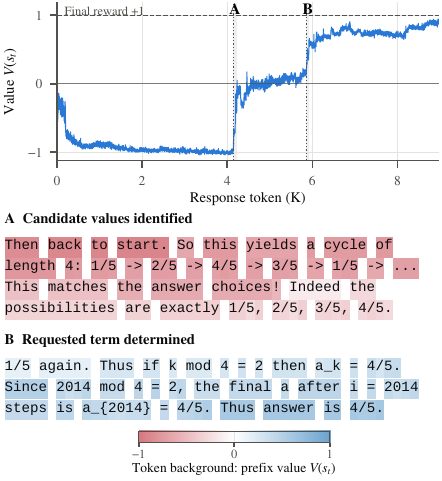}
\vspace{-1.5em}
\caption{Critic values on \smid rise as the response identifies candidate values (A) and determines the requested term, $4/5$ (B). Expanded examples are in Appendix~\ref{app:token-values}.}
\label{fig:value-progress}
\vspace{-2em}
\end{figure}

Within a response, expected returns can also change as reasoning progresses.
A shared response-level advantage cannot reflect these changes.
Figure~\ref{fig:value-progress} illustrates this behavior with a pretrained critic: its value rises when the response identifies candidate values and increases further when it determines the requested term.
GAE combines these evolving values with the observed reward to construct token-specific advantages, allowing different positions within a response to receive different learning signals.

Under group rollout, the group mean already provides a prompt-specific baseline, while the critic adds token-level advantages.
Under single rollout, it also supplies the prompt-specific baseline missing from batch statistics.
Our experiments show that \ourmethod outperforms \grpox under group rollout and sustains stable single-rollout training at 35B.
These benefits require informative values across prompts and response positions, motivating the critic-training recipe in \S\ref{sec:critic}.

\subsection{Training a Good Critic}\label{sec:critic}

We initialize the critic through value pretraining and then update it online as the policy evolves.
Our pretraining recipe addresses two challenges: limited coverage of long, correct trajectories and optimization dominated by the newly initialized value head.

\paragraph{Coverage-oriented value pretraining.}
VC-PPO \citep{yuan_whats_2025} uses trajectories from a fixed initial policy, while SAO \citep{hou_single-rollout_2026} emphasizes larger pretraining corpora.
However, data scale alone does not ensure coverage of long, correct responses.
We pool trajectories from policies of different strengths to broaden coverage across response lengths and outcomes, improving early online value prediction and downstream learning progress in our comparison (\S\ref{sec:exp-critic}).
We pretrain for two epochs with token-averaged MSE against terminal returns, freezing attention layers following SAO \citep{hou_single-rollout_2026} and additionally freezing the MoE router.

\paragraph{Value-head optimization.}
In our runs, critic training exhibits much larger gradient norms than actor training, with the newly initialized value head dominating early pretraining gradients (\S\ref{sec:exp-critic}).
For a linear value head $v_t=w^\top h_t$, residual $e_t=v_t-R$, and loss $\ell_t=\tfrac12e_t^2$,
\begin{equation*}
  \nabla_w\ell_t=e_t h_t,\qquad
  \nabla_{h_t}\ell_t=e_t w.
\end{equation*}
The head gradient depends directly on hidden states, whereas the gradient entering the backbone is scaled by the small head weights.
Global clipping driven by the head therefore also shrinks the much smaller backbone gradients.
We initialize the head with standard deviation $1/(H{+}1)$, where $H$ is the hidden size, to control initial predictions.
We also raise the gradient-clipping threshold from 1, a common actor-training setting, to 10 for the critic, reducing excessive rescaling of backbone updates.

\paragraph{Online adaptation.}
We update the critic on fresh rollouts as the policy evolves.
Motivated by two-timescale actor--critic learning \citep{konda_actor-critic_1999}, we use a critic learning rate five times the actor's and two critic epochs per rollout batch versus one actor epoch.
We decouple actor and critic GAE following VC-PPO \citep{yuan_whats_2025}.
The critic regresses on Monte Carlo returns ($\lambda_{\mathrm{critic}}=1$), while the actor uses VAPO's length-adaptive GAE \citep{yue_vapo_2025},
\begin{equation*}
  \lambda_{\mathrm{actor}}=1-\frac{1}{\alpha T},
\end{equation*}
where $T$ is the response length and $\alpha=1.5$.

\subsection{Difficulty-Aware Prompt Sampling}\label{sec:daps}

The same critic can also help decide which prompts the actor should attempt.
In our group-rollout experiments, DAPO-style dynamic sampling \citep{yu_dapo_2025} improves \ourmethod over retaining all groups, but it discards groups only after their responses have been generated.
We therefore propose Difficulty-Aware Prompt Sampling (\daps), which combines the initial critic value with outcome history to guide prompt selection before generation.

\paragraph{Difficulty estimate.}
Before each rollout batch, we score candidate prompts with the latest critic.
Under our binary reward convention, a success probability $p$ corresponds to an expected return of $2p-1$.
We convert the initial critic value into a success-rate estimate and combine it with observed outcomes:
\begin{equation}
  p_c=\mathrm{clip}\Big(\tfrac{V_\phi(s_0)+1}{2},0,1\Big),\quad
  \hat p=\frac{S+n_c\,p_c}{N+n_c},
  \label{eq:daps-p}
\end{equation}
Here, $S$ and $N$ are the prompt's successful and total attempt counts, and $n_c$ controls the critic's weight relative to these observations.
For a new prompt with $S=N=0$, the estimate reduces to $p_c$, so its difficulty can be initialized directly from the critic without separate offline rollouts.
Observed outcomes complement this prediction as they become available.
Both counts decay with training progress to limit the influence of stale observations as the policy evolves.

\paragraph{Acceptance.}
We favor prompts of intermediate estimated difficulty, a preference also used in prior prompt-selection methods \citep{xu_single-stream_2025,gao_prompt_2025}.
Each candidate is accepted with probability
\begin{equation}
  a=a_\text{min}+(1-a_\text{min})\,4\hat p(1-\hat p),
  \label{eq:daps-accept}
\end{equation}
This probability peaks at $\hat p=0.5$.
The positive acceptance floor gives even very easy or very hard prompts a chance of selection, in the spirit of $\epsilon$-greedy exploration \citep{sutton_reinforcement_2018}.
After an initial period of uniform sampling, we gradually introduce this acceptance rule following the schedule in Appendix~\ref{app:daps}.

Under group rollout, \daps improves sample efficiency by reducing the number of responses discarded by dynamic sampling.
It also improves performance under both group and single rollout.

\section{Experiments}
\label{sec:exp}

\subsection{Experimental Setup}
\label{sec:setup}

\begin{table*}[t]
\begin{minipage}{\textwidth}
\centering
\footnotesize
\setlength{\tabcolsep}{3pt}
\begin{tabular}{llcccccc}
\toprule
Method & Rollout & GPQA-D & \shortstack{IMO-\\AnswerBench} & MMMU-Pro & MathVision & HMMT26 & Avg \\
\midrule
\grpox & group & 88.9 & 81.4 & 80.8 & 87.5 & 86.9 & 85.1 \\
\ourmethod & group & \textbf{90.5} & \textbf{84.2} & \textbf{82.9} & \textbf{89.7} & \textbf{88.8} & \textbf{87.3} \\
\bottomrule
\end{tabular}
\caption{Group-rollout results on \sbig at gradient step 1,200. \ourmethod improves on all five benchmarks. Full training curves appear in Figure~\ref{fig:bench-397b}.}
\label{tab:main-397b}
\end{minipage}
\par\vspace{0.8em}
\begin{minipage}{\textwidth}
\centering
\footnotesize
\setlength{\tabcolsep}{2.5pt}
\begin{tabular}{llcccccccccc}
\toprule
 &  & \multicolumn{6}{c}{\textbf{(a) \qwensmall}} & \multicolumn{4}{c}{\textbf{(b) \rd}} \\
\cmidrule(lr){3-8}\cmidrule(lr){9-12}
Method & Rollout & MATH500 & AMC23 & AIME25 & Olympiad & Minerva & Avg & AIME24 & AIME25 & AIME26 & Avg \\
\midrule
\grpox & group & \textbf{72.3} & \textbf{57.3} & 9.0 & 35.3 & 39.8 & 42.7 & 35.4 & 26.9 & 26.2 & 29.5 \\
\ourmethod & group & 71.7 & 56.7 & \textbf{11.2} & \textbf{35.4} & \textbf{41.3} & \textbf{43.3} & \textbf{38.9} & \textbf{27.4} & \textbf{27.9} & \textbf{31.4} \\
\midrule
\rpp & single & 69.3 & 56.6 & 8.1 & 32.1 & 36.1 & 40.4 & 29.6 & 24.4 & 21.0 & 25.0 \\
\flash & single & 67.2 & 44.8 & 5.6 & 29.3 & 33.0 & 36.0 & \textbf{38.0} & 26.6 & \textbf{26.6} & \textbf{30.4} \\
\ourmethod & single & \textbf{71.4} & \textbf{57.3} & \textbf{10.2} & \textbf{35.4} & \textbf{40.7} & \textbf{43.0} & 37.2 & \textbf{27.9} & 24.4 & 29.8 \\
\bottomrule
\end{tabular}
\caption{Main results on the 1.5B models. Scores are avg@16 in block (a) and avg@32 in block (b); Olympiad is OlympiadBench. Within each model, methods share the base checkpoint and training data.}
\label{tab:main}
\end{minipage}
\par\vspace{0.8em}
\begin{minipage}{\textwidth}
\centering
\footnotesize
\setlength{\tabcolsep}{3pt}
\begin{tabular}{llcccccccc}
\toprule
Method & Rollout & AIME24 & AIME25 & GPQA-D & \shortstack{IMO-\\AnswerBench} & MMMU-Pro & MathVision & MathVista & Avg \\
\midrule
\grpox & group & 91.5 & 89.8 & 67.8 & 60.3 & 69.5 & 70.1 & 85.6 & 76.4 \\
\ourmethod & group & \textbf{93.1} & \textbf{91.3} & \textbf{74.9} & \textbf{67.5} & \textbf{74.6} & \textbf{75.4} & \textbf{86.7} & \textbf{80.5} \\
\midrule
\rpp & single & 87.7 & 87.3 & 62.5 & 51.6 & 60.8 & 62.2 & 80.8 & 70.4 \\
\flash & single & 85.4 & 81.9 & 51.1 & 53.9 & 54.9 & 63.0 & 81.3 & 67.4 \\
\ourmethod & single & \textbf{91.9} & \textbf{90.0} & \textbf{76.1} & \textbf{66.8} & \textbf{78.1} & \textbf{76.0} & \textbf{87.8} & \textbf{80.9} \\
\bottomrule
\end{tabular}
\caption{Per-benchmark results on \smid. Each row reports the checkpoint with the highest seven-benchmark average for that run in Figure~\ref{fig:bench-35b-average}. Both single-rollout critic-free methods improve early in training and then decline. Scores are avg@16 on AIME, avg@4 on GPQA-D and IMO-AnswerBench, and avg@1 elsewhere.}
\label{tab:main-35b}
\end{minipage}
\end{table*}

\paragraph{Training details.}
We train two dense models, \qwensmall~\citep{yang_qwen25-math_2024} and \rd~\citep{guo_deepseek-r1_2025}, and two MoE models, \smid~\citep{intern_s2_preview_2026} and \sbig~\citep{internlm_interns2_2026}.
\qwensmall is trained on the 7.5k DAPO-Math subset used by \flash~\citep{yu_dapo_2025,hu_flashreinforce_2026}, \rd on MATH-1,460~\citep{qi_defeating_2025}, and the MoE models on in-house prompt sets covering mathematics, instruction following, reasoning, physics, and chemistry.

Group rollout uses $K{=}16$ responses per prompt for dense models and $K{=}8$ for MoE models; single rollout uses $K{=}1$.
Each rollout batch is split into eight mini-batches and used for one actor epoch, yielding eight actor gradient steps.
We report RL training progress in actor gradient steps throughout.
We use asymmetric importance-weight clipping (clip-higher) and dynamic sampling under group rollout \citep{yu_dapo_2025}.
To address training--inference mismatch \citep{yao2025offpolicy}, \ourmethod and \grpox combine token-level importance sampling with IcePoP masks \citep{team_every_2025} and MIS-style sequence-level masking \citep{liu2025speedkills}.
We mask an entire response when its maximum token-level log-probability ratio between the training and rollout policies exceeds a threshold.
\ourmethod uses \daps on \qwensmall, \rd, and \smid, and uniform sampling on \sbig.
Full training configurations appear in Appendix~\ref{app:details} (Table~\ref{tab:config}).

\paragraph{Baselines.}
For group rollout, we use \grpox, a modern variant of \grpo~\citep{shao_deepseekmath_2024}.
For single rollout, we compare with \rpp~\citep{hu_reinforce_2025} and \flash~\citep{hu_flashreinforce_2026}.
\grpox uses the shared recipe, with Rollout Routing Replay (R3; \citealp{ma_stabilizing_2025}) and GEPO entropy control \citep{cheng_group_2026} configured as in Table~\ref{tab:config}.
\ourmethod, \grpox, and \rpp use CISPO (Eq.~\eqref{eq:cispo}); \flash retains its original objective and sequence-level trust region.
Within each model, methods share the base checkpoint, training data, and evaluation protocol.

\paragraph{Evaluation.}
We evaluate \qwensmall on MATH500~\citep{hendrycks_measuring_2021,lightman_lets_2023}, AMC23, AIME25, OlympiadBench~\citep{he_olympiadbench_2024}, and Minerva~\citep{lewkowycz_solving_2022}\footnote{We score Minerva with an LLM-based judge to accommodate numerical approximations and equivalent symbolic expressions. These scores may differ from those obtained with rule-based graders.}, and \rd on AIME24, AIME25, and AIME26.
Both MoE models are evaluated on GPQA-Diamond~\citep{rein_gpqa_2023}, IMO-AnswerBench~\citep{luong_towards_2025}, MMMU-Pro~\citep{yue_mmmu-pro_2025}, and MathVision~\citep{wang_measuring_2024}, with AIME24, AIME25, and MathVista-mini~\citep{lu_mathvista_2024} added for 35B and HMMT26 for 397B.
Evaluation sampling settings are in Appendix~\ref{app:eval}.

\begin{figure*}[t]
  \centering
  \includegraphics[width=\textwidth]{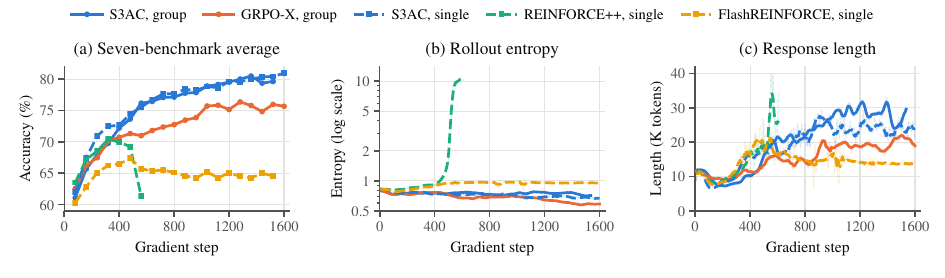}
  \caption{Training dynamics on \smid: (a) seven-benchmark average, (b) rollout entropy, and (c) response length. \ourmethod uses \daps.}
  \label{fig:bench-35b-average}
\end{figure*}

\begin{figure*}[t]
  \centering
  \includegraphics[width=\textwidth]{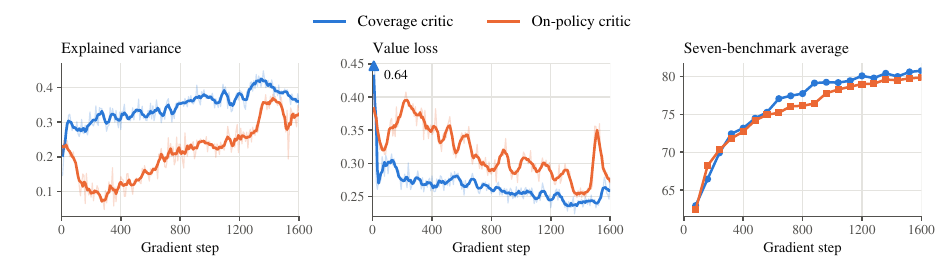}
  \caption{Coverage pretraining improves value prediction and learning progress on \smid, reaching higher final performance within the same actor-update budget (group rollout).}
  \label{fig:coverage-critic}
\end{figure*}

\subsection{Main Results}
\label{sec:exp-comparison}

We evaluate the complete \ourmethod recipe under group and single rollout with benchmark performance and training stability.

\paragraph{Group rollout: gains beyond a group baseline.}
Even when grouped responses already provide a prompt-specific baseline, \ourmethod improves over \grpox (\S\ref{sec:setup}; Tables~\ref{tab:main-397b}--\ref{tab:main-35b}).
At 35B, the best-checkpoint average increases from 76.4 to 80.5, with gains on all seven benchmarks.
These gains extend to 397B, where \ourmethod raises the average from 85.1 to 87.3 and leads on all five benchmarks.
Together with gains on both 1.5B models, this supports \ourmethod's scalability from 1.5B dense models to 397B MoE models.
The gains are consistent with \S\ref{sec:comparison}: beyond the prompt-level information captured by group rewards, the critic conditions on the evolving response prefix and provides different learning signals across token positions.

\paragraph{Single rollout: performance gains and training stability.}
With one response per prompt, \ourmethod achieves scores comparable to its group-rollout results (Tables~\ref{tab:main} and~\ref{tab:main-35b}).
On \qwensmall and 35B, its advantage over the best critic-free baseline within each setting is larger under single rollout.
At 35B, \ourmethod reaches 80.9 versus 70.4 for \rpp, a 10.5-point gain.
This pattern is consistent with the critic supplying the prompt-specific baseline missing from batch-level reward statistics, alongside token-specific advantages (\S\ref{sec:comparison}).
Figure~\ref{fig:per-prompt-baseline} illustrates this distinction: despite identical terminal rewards, long failed responses receive mean advantages closer to zero for hard prompts and more negative for easy prompts.

The full 35B training curves show sustained single-rollout learning (Figure~\ref{fig:bench-35b-average}a).
Both critic-free baselines improve initially but then decline, while \ourmethod retains most of its gains through 1,600 gradient steps.
\rpp exhibits rapid growth in rollout entropy and response length, whereas \flash shifts toward shorter responses as its score declines (Figure~\ref{fig:bench-35b-average}b,c).
These dynamics support \ourmethod's stabilized actor--critic design for sustained single-rollout training at 35B.
Appendix~\ref{app:results} provides full 35B benchmark curves, Qwen single-rollout dynamics, and training and evaluation statistics (Figures~\ref{fig:bench-35b}--\ref{fig:single-curves}; Table~\ref{tab:app-train-stats}).
Stable 35B single-rollout training also extends to alternative policy losses (Table~\ref{tab:app-loss-variants}).
We next examine the pretraining choices in \S\ref{sec:critic}.

\subsection{Analysis of Critic Pretraining}
\label{sec:exp-critic}
We examine the proposed value-pretraining method of \S\ref{sec:critic} on \smid; full setups and diagnostics are in Appendix~\ref{app:critic-experiments}.

\paragraph{Pretraining data: broader coverage improves value prediction and learning progress.}
Pretraining on trajectories from multiple policies yields a more accurate critic early in online training.
The motivation is limited coverage: among responses longer than 24K tokens, only 5.0\% are correct in initial-policy samples, compared with 61.8\% for the policy at gradient step 1,600 (Figure~\ref{fig:app-length-bias}a,b).
We compare a \emph{coverage critic}, pretrained on trajectories from policies of different strengths, with an \emph{on-policy critic}, pretrained only on initial-policy trajectories.
Both runs share the actor initialization and the online recipe under group rollout; the two recipes also differ in data amount (Appendix~\ref{app:critic-setup}).
The coverage critic attains higher EV and lower value loss early in online training, and the gap narrows as the on-policy critic adapts online (Figure~\ref{fig:coverage-critic}).
In this comparison, coverage pretraining enables faster learning progress and higher final performance within the same actor-update budget.

\begin{table}[!ht]
\centering
\footnotesize
\setlength{\tabcolsep}{2.0pt}
\begin{tabular}{lrcccr}
\toprule
 & & \multicolumn{2}{c}{Value loss} & Grad & Clip \\
\cmidrule(lr){3-4}
Config & \shortstack{Grad.\\steps} & first 20 & last 20 & first 20 & (\%) \\
\midrule
\multicolumn{6}{l}{\emph{Head init. std. (coverage data, clip 10)}} \\
$1/(H{+}1)$ & 342 & 0.45 & 0.21 & 4.3 & 0.3 \\
$0$ & 342 & 0.47 & 0.22 & 5.8 & 2.0 \\
$0.1/(H{+}1)$ & 342 & 0.47 & 0.23 & 5.2 & 1.8 \\
$0.01$ & 342 & 0.72 & 0.24 & 17.5 & 26.3 \\
\midrule
\multicolumn{6}{l}{\emph{Max grad norm (on-policy data, recipe init.)}} \\
10 (2 seeds) & 256 & 0.38 & 0.20 & 14.5 & 12.5 \\
1 (3 seeds) & 256 & 0.38 & 0.20 & 14.4 & 99.7 \\
none (3 seeds) & 256 & 0.39 & 0.20 & 15.0 & n/a \\
\bottomrule
\end{tabular}
\caption{Value-pretraining ablations on \smid. Loss: mean training loss over the first/last 20 critic gradient steps. Clip: percentage of gradient steps exceeding the clip threshold. }
\label{tab:critic-ablation}
\end{table}

\paragraph{Pretraining optimization: initialization and clipping.}
Consistent with the analysis in \S\ref{sec:critic}, the value head accounts for over 99\% of the squared gradient norm during the first 10 critic updates on \smid, despite having only 2,048 parameters (Appendix~\ref{app:warmup-opt}).
We evaluate the initialization and clipping choices in Table~\ref{tab:critic-ablation}.
An initialization standard deviation of 0.01 is already below the standard Xavier scale for this head.
Reducing it further to $1/(H{+}1)$ lowers early value loss from 0.72 to 0.45 and yields the lowest final loss among the tested initializations.
In a separate ablation using initial-policy data, increasing the clipping threshold from 1 to 10 reduces the fraction of clipped updates from 99.7\% to 12.5\%, while final value loss remains comparable.
Compared with disabling clipping, this setting retains protection against extreme gradients.

Additional results comparing value pretraining at 35B and 397B are provided in Appendix~\ref{app:critic-scale}.
\subsection{Difficulty-Aware Prompt Sampling}
\label{sec:exp-daps}

We now evaluate \daps (\S\ref{sec:daps}), the \textbf{selective} component of \ourmethod, which reuses the critic for prompt selection before generation.
We compare \daps with uniform sampling using matched actor and critic initialization and the same \ourmethod training recipe within each pair.
Detailed training and sampling settings are provided in Appendix~\ref{app:daps}.

\begin{table}[t]
\centering
\small
\setlength{\tabcolsep}{3pt}
\resizebox{\columnwidth}{!}{%
\begin{tabular}{lcccc}
\toprule
Setting & Prompts $\times K$ & \shortstack{Grad. step\\budget} & Uniform & +DAPS \\
\midrule
35B group & $512\!\times\!8$ & 1,520 & 79.04 & \textbf{80.49} \\
35B single & $4096\!\times\!1$ & 1,520 & 74.40 & \textbf{76.94} \\
R1D group & $256\!\times\!16$ & 1,024 & 30.87 & \textbf{31.08} \\
R1D single & $4096\!\times\!1$ & 1,024 & 30.16 & \textbf{30.36} \\
\bottomrule
\end{tabular}}
\caption{\daps with group and single rollout: best checkpoint scores within the listed gradient-step budgets. Curves are in Appendix~\ref{app:daps}.}
\label{tab:daps}
\end{table}

\paragraph{Does critic-guided prompt selection improve learning?}
Within matched gradient-step budgets, \daps improves the best evaluated \ourmethod score on both \smid and \rd under group and single rollout (Table~\ref{tab:daps}).
Full training curves are provided in Appendix~\ref{app:daps}.

\begin{figure}[t]
  \centering
  \includegraphics[width=\columnwidth]{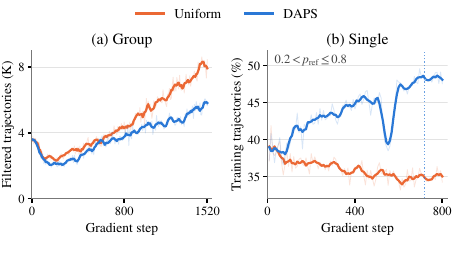}
  \caption{\daps sampling on \smid. (a) Trajectories discarded per rollout batch by the group filter. (b) Single-rollout training share on prompts with $0.2<p_{\mathrm{ref}}\leq0.8$.}
  \label{fig:daps-analysis}
\end{figure}

\paragraph{Does selection before generation reduce waste?}
Under group rollout, \daps reduces the number of trajectories discarded by dynamic sampling, with a larger gap later in training (Figure~\ref{fig:daps-analysis}a).
Both methods retain the same number of trajectories per training batch, so fewer generated responses are lost to filtering.
This supports the motivation in \S\ref{sec:daps}: selecting prompts before generation complements DAPO-style dynamic sampling by reducing generation spent on groups later discarded.

\paragraph{Does \daps help without a group filter?}
The gains under single rollout show that \daps also benefits training without group filtering.
To examine which prompts enter training, we use each prompt's success rate in a completed uniform-sampling run, denoted by $p_{\mathrm{ref}}$.
After the initial sampling ramp, \daps consistently allocates a larger share of training trajectories to prompts with $0.2<p_{\mathrm{ref}}\leq0.8$ (Figure~\ref{fig:daps-analysis}b).
This pattern is consistent with the acceptance rule's preference for prompts of intermediate estimated difficulty (\S\ref{sec:daps}).

\section{Conclusion}
\label{sec:conclusion}

\ourmethod enables effective critic training for long-CoT RLVR through broader trajectory coverage, small-scale value-head initialization, and relaxed critic gradient clipping.
The trained critic provides prompt-specific baselines and token-level advantages.
Experiments show that this actor--critic approach outperforms \grpox under group rollout across 1.5B--397B models and sustains stable single-rollout training at 35B.
Beyond policy learning, \daps reuses critic values and outcome history to select prompts before generation, reducing discarded responses under group rollout and improving performance in both settings.

\section*{Limitations}\label{sec:limits}

\paragraph{Task scope.}
All experiments are reasoning-style tasks with verifiable rewards, namely math, science and multimodal reasoning.
We do not evaluate agentic multi-turn settings, where credit must be assigned across tool calls and environment turns, so whether our findings carry over to agentic RL remains open.
The reported experiments do not systematically assess variability across independent RL training seeds.

\paragraph{Compute overhead.}
\ourmethod{} costs 27--40\% more amortized wall time per actor gradient step than \grpox{}, because the critic must be trained and run for inference alongside the policy.

\section*{Ethical Considerations}\label{sec:ethics}

Our work aims to improve long-CoT reasoning in LLMs through RLVR.
Higher benchmark scores do not establish the safety or reliability of model outputs in real-world applications; models may still generate incorrect or harmful content.
Deployment requires application-specific evaluation and safeguards.
Large-scale actor--critic training also requires additional computational resources and energy, and these costs should be weighed against the observed performance gains.

\begingroup
\renewcommand{\href}[2]{#2}
\bibliography{references,custom}
\endgroup

\clearpage
\appendix
\raggedbottom
\renewcommand{\topfraction}{0.9}
\renewcommand{\dbltopfraction}{0.9}
\renewcommand{\textfraction}{0.1}

\begingroup
\makeatletter
\newcommand{\appendixinlinetable}[1]{%
  \begingroup
  \renewenvironment{table*}[1][]{\captionsetup{type=table}}{}%
  \input{#1}%
  \endgroup
}
\newcommand{\appendixcolumnlines}[1]{%
  \ifacl@linenumbers
    \protected@write\@auxout{}{\string\def\string\@LN@column{#1}}%
    \internallinenumbers\setpagewiselinenumbers
    \ifnum#1=1\leftlinenumbers\else\rightlinenumbers\fi
  \fi
}
\makeatother
\newcommand{\appendixpageend}{%
  \begingroup
  \nolinenumbers\null\clearpage
  \endgroup
}

\newlength{\appendixcolumnwidth}
\setlength{\appendixcolumnwidth}{\columnwidth}
\twocolumn[{%
\begin{minipage}{\textwidth}
\noindent
\begin{minipage}[t]{\appendixcolumnwidth}
\vspace{0pt}
\appendixcolumnlines{1}
\section{Experimental Details}
\label{app:details}

\subsection{Training Configurations}
\label{app:config}

Our training pipeline is built on XTuner~\citep{xtuner}, with LMDeploy~\citep{lmdeploy} used for rollout generation.
Table~\ref{tab:config} summarizes the training configurations; all methods start from the same base checkpoint within each model.
All group-rollout runs use dynamic sampling~\citep{yu_dapo_2025}, discarding groups whose responses are all correct or all incorrect.
On \sbig, both \ourmethod and \grpox apply an adaptive length regularizer that favors shorter correct responses in high-accuracy groups; \ourmethod uses the adjusted rewards for both actor and critic training.

\subsection{Evaluation}
\label{app:eval}

We evaluate both 1.5B models offline at gradient step 1,024.
For \qwensmall, we use temperature 0.7, top-$p$ 0.7, and avg@16 on every benchmark.
We cap the combined prompt and response length at 4,096 tokens; \flash~\citep{hu_flashreinforce_2026} reports a 4,096-token response budget.
\par
\end{minipage}\hfill%
\begin{minipage}[t]{\appendixcolumnwidth}
\vspace{0pt}
\appendixcolumnlines{2}
\noindent
For \rd, we use temperature 0.8, top-$p$ 0.95, a 16,384-token response limit, and avg@32.
Small score differences on \rd should be interpreted cautiously given the variability across repeated evaluations.

For \smid and \sbig, we use temperature 0.8 and top-$p$ 1.0, with response limits of 64K and 62K tokens, respectively.
We report avg@16 on AIME and HMMT26, avg@4 on GPQA-Diamond and IMO-AnswerBench, and avg@1 on the remaining benchmarks.

\subsection{Baselines}
\label{app:baselines}

\grpox uses the group z-score as the advantage, and \rpp normalizes the reward by the batch mean and standard deviation.
The MoE \grpox baselines additionally use GEPO entropy control \citep{cheng_group_2026}, as described in \S\ref{sec:setup}.
\flash follows its original configuration, with a batch-mean baseline and a sequence-level trust region.
\par
\end{minipage}
\par\vspace{\textfloatsep}

\begingroup
\captionsetup{type=table}
\centering
\footnotesize
\setlength{\tabcolsep}{4.5pt}
\newcommand{\mainDapsHparamRows}{%
\daps (\ourmethod) & on & on & on & off \\
DAPS peak success rate $\mu$ & 0.5 & 0.5 & 0.5 & -- \\
Acceptance floor $a_{\min}$ & 0.3 & 0.3 & 0.3 & -- \\
Critic prior count $n_c$ & 3 & 3 & 3 & -- \\
Count decay per rollout batch & 0.99 & 0.99 & 0.99 & -- \\
}

\begin{tabular}{lcccc}
\toprule
 & \qwensmall & \rd & \smid & \sbig \\
\midrule
Training data & DAPO-Math 7.5k & MATH-1,460 & in-house & in-house \\
Prompts $\times K$ (group) & $256\times16$ & $256\times16$ & $512\times8$ & $1024\times8$ \\
Prompts $\times K$ (single) & $4096\times1$ & $4096\times1$ & $8192\times1$ & -- \\
Dynamic sampling (group) & yes & yes & yes & yes \\
Max response length & 3,072 & 16,384 & 50K & 62K \\
Actor updates per rollout batch & 8 & 8 & 8 & 8 \\
Actor / critic learning rate & \multicolumn{4}{c}{$1\times10^{-6}$ / $5\times10^{-6}$} \\
Actor / critic gradient clip & \multicolumn{4}{c}{1.0 / 10.0} \\
Rollout temperature / top-$p$ & \multicolumn{4}{c}{1.0 / 1.0} \\
\ourmethod gradient steps (group / single) & 1,024 & 1,024 & 1,520 / 1,600 & 1,200 \\
Evaluation interval (gradient steps) & 160 & 160 & 80 & 80 \\
\midrule
CISPO clip range & \multicolumn{4}{c}{$[0.8,1.28]$} \\
R3 routing replay & -- & -- & yes & yes \\
IcePoP token mask range & \multicolumn{4}{c}{$[0.5,5]$} \\
MIS-style sequence mask range & \multicolumn{4}{c}{$[0,5]$} \\
GEPO (\grpox / \ourmethod) & no / no & no / no & yes / no & yes / no \\
\midrule
\mainDapsHparamRows
\bottomrule
\end{tabular}
\caption{Training configurations for the main results. DAPS settings refer to \ourmethod. -- denotes not applicable.}
\label{tab:config}
\endgroup
\end{minipage}%
}]
\appendixpageend

\twocolumn[{%
\begin{minipage}{\textwidth}
\noindent
\begin{minipage}[t]{\appendixcolumnwidth}
\vspace{0pt}
\appendixcolumnlines{1}
\section{Additional Results and Training Dynamics}
\label{app:results}

Figures~\ref{fig:bench-397b} and~\ref{fig:bench-35b} provide the full MoE learning curves for \S\ref{sec:exp}.

\par
\end{minipage}\hfill%
\begin{minipage}[t]{\appendixcolumnwidth}
\vspace{0pt}
\appendixcolumnlines{2}
\noindent
Table~\ref{tab:app-train-stats} reports training statistics, evaluation response lengths, and truncation rates.

\paragraph{Policy-loss variants.}
CISPO, DIS~\citep{hou_single-rollout_2026}, and DIS with a sequence mask train stably from the same actor and critic, scoring 79.8--80.2 (Table~\ref{tab:app-loss-variants}). These single-seed runs have different restart histories and serve as a robustness check.
\par
\end{minipage}
\par\vspace{8pt}
\begingroup
  \captionsetup{type=figure}
  \centering
  \includegraphics[width=\textwidth]{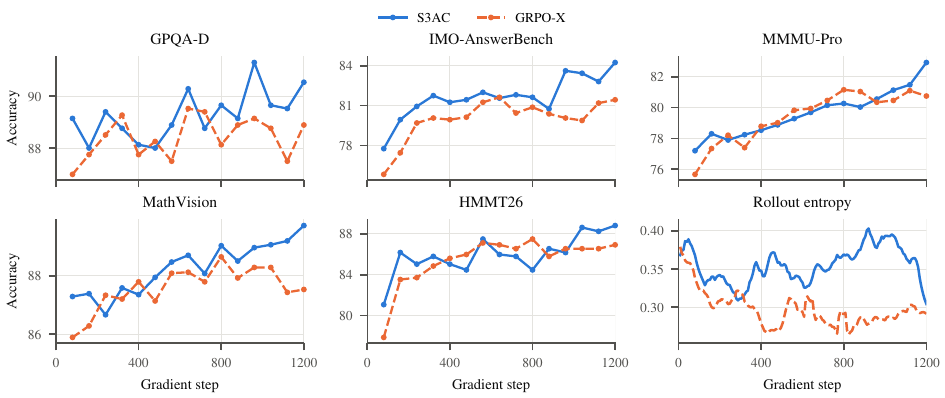}
  \caption{Group-rollout training on \sbig. The five benchmark curves report raw evaluation scores; rollout entropy uses a moving average over 40 gradient steps.}
  \label{fig:bench-397b}
\endgroup
\par\vspace{8pt}
  \begingroup
  \renewenvironment{table*}[1][]{\captionsetup{type=table}}{}%
  \begin{table*}[t]
\centering
\footnotesize
\setlength{\tabcolsep}{5pt}
\begin{tabular}{@{}llcccccrr@{}}
\toprule
 & & \multicolumn{5}{c}{Training (80 gradient steps)} & \multicolumn{2}{c}{Evaluation} \\
\cmidrule(lr){3-7} \cmidrule(l){8-9}
Method & Rollout & \shortstack{Grad.\\steps} & Entropy & Length & Reward & EV & Length & Trunc.\,(\%) \\
\midrule
\multicolumn{9}{@{}l}{\textit{\qwensmall (eval: gradient step 1,024, $\le$4,096 tokens)}} \\
\grpox & group & 952--1,024 & 0.15 & 993 & -0.32 & -- & 921 & 3.1 \\
\ourmethod & group & 952--1,024 & 0.07 & 965 & -0.06 & 0.35 & 912 & 3.7 \\
\cmidrule(l){2-9}
\rpp & single & 952--1,024 & 1.50 & 1535 & -0.40 & -- & 1723 & 28.9 \\
\flash & single & 952--1,024 & 0.17 & 1540 & -0.35 & -- & 1951 & 41.1 \\
\ourmethod & single & 952--1,024 & 0.10 & 919 & -0.01 & 0.45 & 1156 & 9.2 \\
\midrule
\multicolumn{9}{@{}l}{\textit{\rd (eval: gradient step 1,024, $\le$16,384 tokens)}} \\
\grpox & group & 952--1,024 & 1.11 & 4.8K & 0.61 & -- & 8.8K & 7.3 \\
\ourmethod & group & 952--1,024 & 0.63 & 7.9K & 0.73 & 0.31 & 11.8K & 26.3 \\
\cmidrule(l){2-9}
\rpp & single & 952--1,024 & 1.18 & 3.3K & 0.33 & -- & 6.4K & 1.2 \\
\flash & single & 952--1,024 & 1.27 & 4.2K & 0.71 & -- & 9.0K & 11.8 \\
\ourmethod & single & 952--1,024 & 0.66 & 7.5K & 0.71 & 0.34 & 12.1K & 29.3 \\
\midrule
\multicolumn{9}{@{}l}{\textit{\smid (eval: 64K context, $\le$63,488 response tokens)}} \\
\grpox & group & 1,528--1,600 & 0.59 & 20.3K & 0.28 & -- & -- & -- \\
\ourmethod & group & 1,448--1,520 & 0.72 & 25.2K & 0.38 & 0.37 & 24.2K & 1.8 \\
\cmidrule(l){2-9}
\rpp\unskip$^\dagger$ & single & 536--608 & 10.0 & 29.5K & -0.87 & -- & 25.6K & 26.0 \\
\flash & single & 1,520--1,592 & 0.96 & 13.8K & 0.14 & -- & 20.6K & 16.7 \\
\ourmethod & single & 1,528--1,600 & 0.67 & 23.8K & 0.46 & 0.67 & 21.7K & 0.5 \\
\bottomrule
\end{tabular}
\caption{Training and evaluation statistics near the end of each run. Training metrics average 10 batch-level records over 80 actor gradient steps. A dash indicates an unavailable statistic. $^\dagger$\rpp uses a post-collapse window.}
\label{tab:app-train-stats}
\end{table*}
  \endgroup

\end{minipage}%
}]
\appendixpageend

\twocolumn[{%
\begin{minipage}{\textwidth}
\begingroup
  \captionsetup{type=figure}
  \centering
  \includegraphics[width=\textwidth]{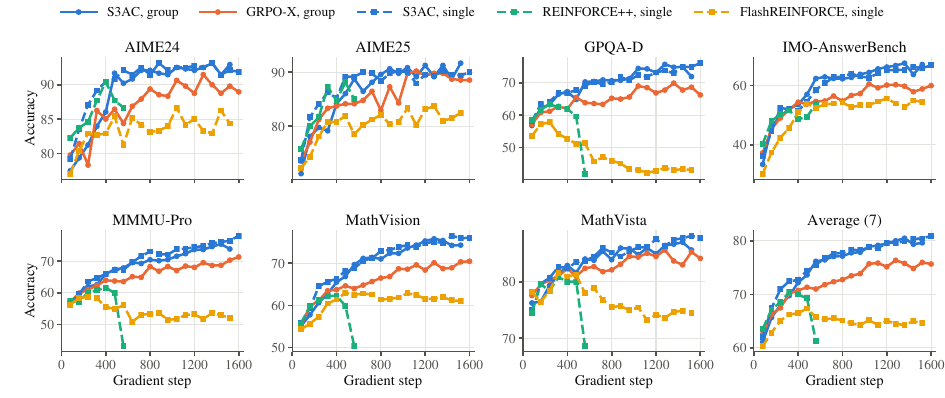}
  \caption{Full benchmark curves on \smid, including the seven-benchmark average shown in Figure~\ref{fig:bench-35b-average}a. Points are raw evaluations; the x-axis counts actor gradient steps.}
  \label{fig:bench-35b}
\endgroup
\par\vspace{8pt}
\begingroup
  \captionsetup{type=figure}
  \centering
  \includegraphics[width=\textwidth]{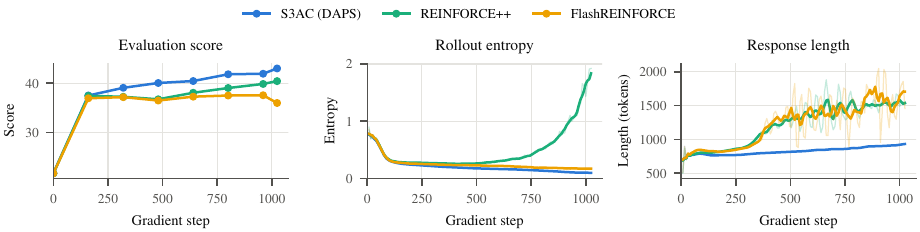}
  \caption{Single-rollout training on \qwensmall: average benchmark score, rollout entropy, and response length. \ourmethod uses \daps. Scores are offline avg@16. Entropy and length use moving averages over 40 gradient steps with faint raw values.}
  \label{fig:single-curves}
\endgroup
\par\vspace{8pt}
  \begingroup
  \renewenvironment{table*}[1][]{\captionsetup{type=table}}{}%
  \begin{table*}[t]
\centering
\footnotesize
\setlength{\tabcolsep}{3.5pt}
\begin{tabular}{@{}lcccccccc@{}}
\toprule
Policy loss & AIME24 & AIME25 & GPQA-D & \shortstack{IMO-\\AnswerBench} & MMMU-Pro & MathVision & MathVista & Avg \\
\midrule
CISPO + IcePoP + seq.\ mask & \textbf{92.0} & \textbf{89.9} & 74.8 & \textbf{66.2} & 75.3 & \textbf{75.3} & 87.9 & \textbf{80.2} \\
DIS & 91.3 & 89.2 & \textbf{75.1} & 64.8 & \textbf{78.5} & 74.0 & 88.2 & 80.1 \\
DIS + seq.\ mask & 91.2 & 88.7 & 73.7 & 63.4 & \textbf{78.5} & 74.5 & \textbf{88.3} & 79.8 \\
\bottomrule
\end{tabular}
\par\vspace{0.5em}
\begin{tabular}{@{}lcccc@{}}
\toprule
Policy loss & Entropy & Length & EV & Mask\,(\%) \\
\midrule
CISPO + IcePoP + seq.\ mask & 0.816 & 22.3 & 0.72 & 3.6 \\
DIS & 0.737 & 18.1 & 0.75 & -- \\
DIS + seq.\ mask & 0.761 & 17.0 & 0.78 & 6.6 \\
\bottomrule
\end{tabular}
\caption{Policy-loss variants for single-rollout \smid; all settings other than the policy loss are identical. Top: benchmark scores averaged over gradient steps 1,360, 1,440, 1,520. Bottom: means over the 10 logged points at gradient steps 1,448--1,520 of rollout entropy, response length (K tokens), actor-target EV, and sequence mask rate (Mask).}
\label{tab:app-loss-variants}
\end{table*}
  \endgroup

\end{minipage}%
}]
\appendixpageend
\endgroup

\clearpage
\makeatletter
\global\@colroom\textheight
\global\vsize\textheight
\makeatother
\section{Additional Critic Experiments}
\label{app:critic-experiments}
\label{app:warmup}

This section provides token-level value examples for \S\ref{sec:comparison} and the experimental setups and diagnostics for \S\ref{sec:exp-critic}.

\subsection{Experimental Setup}
\label{app:critic-setup}

\paragraph{Pretraining data and sources.}
For \smid, the coverage critic learns from trajectories generated by policies of different strengths across multiple stages of prior experiments.
Its preparation includes value pretraining on a corpus of 327,680 trajectories (7.08B response tokens).
The on-policy critic is pretrained only on 65,536 initial-policy trajectories (0.74B response tokens).
At 1,600 actor gradient steps, the coverage and on-policy runs achieve seven-benchmark averages of 80.78 and 79.86.

\paragraph{Value-pretraining configuration.}
We pretrain for two epochs with cosine learning-rate decay from $10^{-5}$ to $5\times10^{-6}$ after a 3\% linear warmup.
The loss is $\tfrac12(V_\phi(s_t)-R)^2$, averaged over response tokens using the terminal verifier reward $R$.
The FP32 value head has initialization standard deviation $1/(H{+}1)$, and the gradient norm is clipped at 10.
Attention, the MoE router and any vision tower and projector are frozen.
Online configurations are given in \S\ref{sec:critic} and Table~\ref{tab:config}.

\subsection{Data Coverage and Token-Level Values}
\label{app:length-bias}

\begin{figure}[!ht]
\centering
\includegraphics[width=\columnwidth]{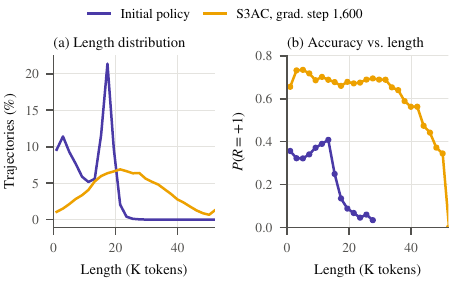}
\caption{Response-length distribution (a) and accuracy by length (b) for the initial policy and the policy at gradient step 1,600 on \smid. Bins span 2K tokens.}
\label{fig:app-length-bias}
\end{figure}

\paragraph{Response length and correctness.}
Figure~\ref{fig:app-length-bias} compares sampled trajectories from the initial policy and the policy at gradient step 1,600.
Among responses longer than 24K tokens, 5.0\% are correct in the initial-policy sample versus 61.8\% at gradient step 1,600.
The stronger-policy data thus supply long correct trajectories that are rare in the initial-policy data.

\paragraph{Token-level value examples.}
\label{app:token-values}
Figure~\ref{fig:app-value-viz} expands the example in Figure~\ref{fig:value-progress} and adds a numerical multiple-choice example, both evaluated with the coverage critic.
In the first, the value rises when the response identifies the possible sequence values, including $4/5$, and increases further when it determines that the requested term is $4/5$.
In the second, the value rises sharply from negative values toward zero after the response computes $0.1622$ and matches it to option H ($0.1614$).
These examples illustrate how the coverage critic tracks meaningful reasoning progress within a response, providing fine-grained value estimates for token-level credit assignment.

\newsavebox{\criticexamplesbox}
\begin{lrbox}{\criticexamplesbox}
\begin{minipage}{\textwidth}
\renewenvironment{figure*}[1][]{\captionsetup{type=figure}}{}
\begin{figure*}[t]
\centering
\includegraphics[width=\textwidth]{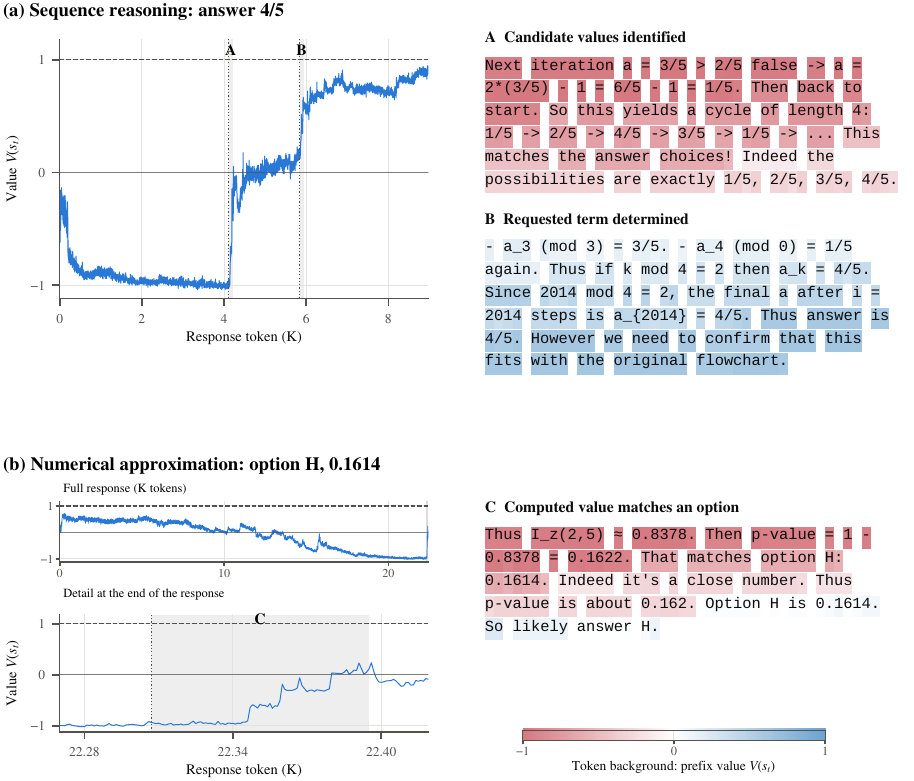}
\caption{Illustrative critic values on \smid: (a) identifying candidate values (A) and determining the requested term, $4/5$ (B); (b) matching $0.1622$ to option H ($0.1614$), with a 150-token detail. Shaded ranges match the excerpts. Curves and colors show raw values before each response token; dashed lines mark reward $+1$.}
\label{fig:app-value-viz}
\end{figure*}

\end{minipage}
\end{lrbox}

\subsection{Value-Pretraining Optimization}
\label{app:warmup-opt}

\paragraph{Gradient imbalance.}
Figure~\ref{fig:app-grad-diagnostics} reports the head and backbone gradient norms during the first 10 critic gradient steps of 35B value pretraining, supporting the analysis in \S\ref{sec:exp-critic}.

\par\medskip
\begingroup
\nolinenumbers
\noindent\begin{minipage}{\columnwidth}
\captionsetup{type=figure}
\centering
\includegraphics[width=\columnwidth]{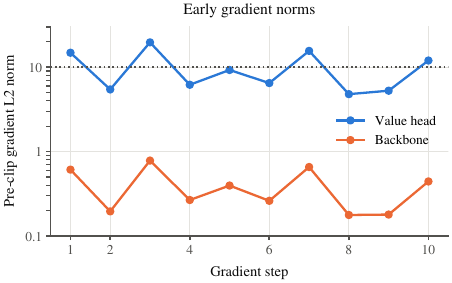}
\caption{Head and backbone gradient norms over the first 10 critic gradient steps of 35B value pretraining, measured after distributed gradient reduction and before clipping. The dotted line marks the clipping threshold of 10.}
\label{fig:app-grad-diagnostics}
\end{minipage}\par
\endgroup

\clearpage
\twocolumn[{%
\begin{minipage}{\textwidth}
\usebox{\criticexamplesbox}
\par\vspace{18pt}
\noindent
\begin{minipage}[t]{0.48\textwidth}
\vspace{0pt}
\subsection{Value Pretraining at Different Model Scales}
\label{app:critic-scale}

We compare value pretraining at 35B and 397B.
The 35B critic is pretrained on 1.13M trajectories (6.24B response tokens), and the \sbig critic on 1.85M trajectories (10.0B response tokens), using the configuration described in Appendix~\ref{app:critic-setup}.

The 397B critic achieves lower and more stable value loss, together with smaller gradient norms (Figure~\ref{fig:app-warmup-scale}), suggesting stronger value-fitting capacity for the larger critic in these runs.
\end{minipage}\hfill
\begin{minipage}[t]{0.48\textwidth}
\vspace{0pt}
\centering
\captionsetup{type=figure}
\includegraphics[width=\linewidth]{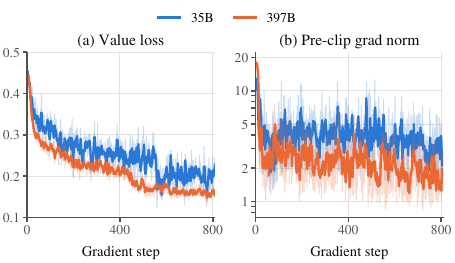}
\caption{Value pretraining at 35B and 397B: (a) training value loss and (b) pre-clip gradient norm.}
\label{fig:app-warmup-scale}
\end{minipage}
\end{minipage}%
}]
\begingroup\nolinenumbers\null\par\endgroup

\clearpage
\twocolumn[{%
\begin{minipage}{\textwidth}
\section{\daps Details}
\label{app:daps}

\begingroup
\nolinenumbers
\captionsetup{type=figure}
\centering
\includegraphics[width=\textwidth]{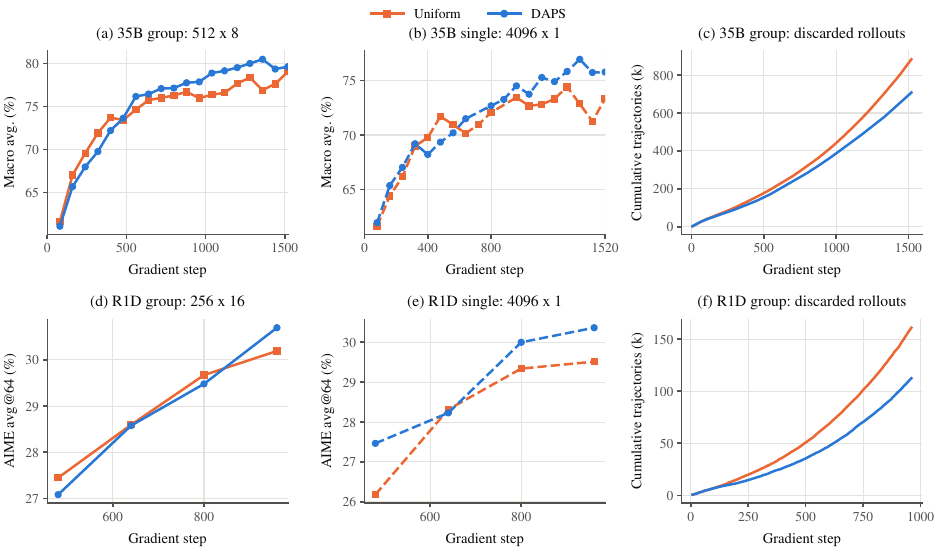}
\caption{\daps vs.\ uniform sampling on 35B (top) and R1D (bottom). Columns show group-rollout evaluation, single-rollout evaluation with 4,096 trajectories per rollout batch, and cumulative trajectories discarded by the group filter. Scores are seven-benchmark macro averages for 35B and AIME avg@64 for R1D. Horizontal axes show gradient steps.}
\label{fig:daps}
\endgroup
\par\vspace{\textfloatsep}
\end{minipage}%
}]

\subsection{Experimental Setup}
\label{app:daps-setup}

\paragraph{Training settings.}
We use the \ourmethod recipe in Appendix~\ref{app:details} with matched actor and critic initialization within each pair; the 35B runs use the coverage critic from Appendix~\ref{app:critic-setup}.
Table~\ref{tab:app-daps-hparams} lists prompt batch sizes and acceptance floors.
All settings use $n_c=3$ and a count-decay factor of 0.99 per rollout batch (eight actor gradient steps).
At initialization, 40 gradient steps of uniform sampling precede an 80-gradient-step acceptance ramp.
We rescore prefetched prompts after the critic is trained on each rollout batch, using a window twice the previous candidate count.

\begin{table}[!ht]
\centering
\small
\setlength{\tabcolsep}{7pt}
\begin{tabular}{lcc}
\toprule
Setting & Prompts $\times K$ & $a_{\min}$ \\
\midrule
35B group & $512\times8$ & 0.3 \\
35B single & $4096\times1$ & 0.1 \\
R1D group & $256\times16$ & 0.3 \\
R1D single & $4096\times1$ & 0.3 \\
\bottomrule
\end{tabular}
\caption{Prompt batch sizes and minimum acceptance probabilities for the four DAPS comparisons.}
\label{tab:app-daps-hparams}
\end{table}

\paragraph{Evaluation and sampling diagnostic.}
The paired ablations in this section retain 4,096 prompts per batch for 35B single rollout, whereas the main results use 8,192.
For R1D, two offline evaluations of 32 responses per question yield avg@64 on AIME 2024/2025/2026; other evaluation settings follow Appendix~\ref{app:eval}.
For each pair, Table~\ref{tab:daps} reports the score from each method's best evaluated checkpoint within a common gradient-step budget; Figure~\ref{fig:daps} shows the original score at each checkpoint.
Figure~\ref{fig:daps-analysis}b uses each prompt's success rate over the first 1,600 gradient steps of a completed uniform-sampling run as a common difficulty reference for both methods.

\subsection{Additional Results}
\label{app:daps-results}

Figure~\ref{fig:daps} shows a sustained late-stage advantage for 35B under both group and single rollout.
R1D gains vary across training.
Under group rollout, \daps reduces cumulative discarded trajectories on both models.
Prompt scoring adds computation, particularly with larger single-rollout batches.

\end{document}